%% file: camera_ready.tex
\documentclass[10pt,twocolumn,letterpaper]{article}

\usepackage{cvpr}              

\usepackage{graphicx}
\usepackage{amsmath}
\usepackage{amssymb}
\usepackage{booktabs}
\usepackage{color}
\usepackage{multirow}
\usepackage{colortbl}
\usepackage{xcolor}
\usepackage[misc]{ifsym}

\definecolor{linkcolor}{RGB}{255,0,0}
\definecolor{urlcolor}{RGB}{255,105,180}
\definecolor{citecolor}{RGB}{66,168,235}
\usepackage[pagebackref,breaklinks,colorlinks]{hyperref}
\hypersetup{colorlinks=true,linkcolor=blue,urlcolor=blue,citecolor=blue}

\def\paperID{1725} 
\def\confName{CVPR}
\def\confYear{2024}

\title{OMG-Seg \raisebox{-0.22\height}{\includegraphics[width=0.05\linewidth]{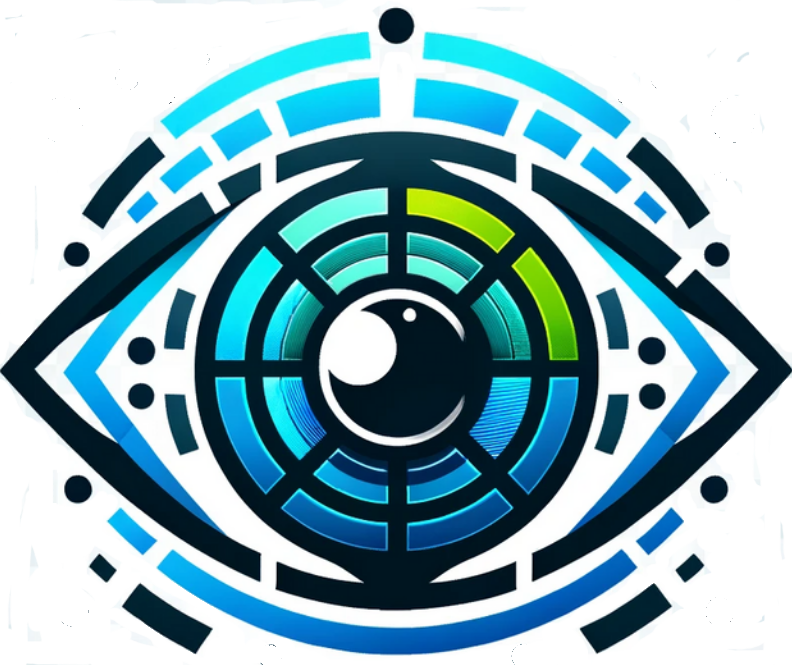}}: Is One Model Good Enough For All Segmentation?}

\author{
     Xiangtai Li$^{1}$ \textsuperscript{$\dagger$} \quad Haobo Yuan$^{1}$   \quad Wei Li$^{1}$ \quad Henghui Ding$^{1}$ \quad Size Wu$^{1}$ \quad Wenwei Zhang$^{1}$ \quad  \\ Yining Li$^{2}$ \quad Kai Chen$^{2}$ \quad Chen Change Loy$^{1}$  \vspace{0.3em} \\
     {\normalsize $^1$S-Lab, Nanyang Technological University \quad $^2$Shanghai Artificial Intelligence Laboratory} \\
    {\normalsize Project Page: \url{https://lxtgh.github.io/project/omg_seg}} \\
    {\normalsize \textsuperscript{$\dagger$}: Project lead and corresponding author. E-mail: xiangtai94@gmail.com \quad ccloy@ntu.edu.sg}
}

\begin{document}
\twocolumn[{%
\renewcommand\twocolumn[1][]{#1}%
\maketitle
\vspace{-8mm}
\centering
\includegraphics[width=0.82\textwidth]{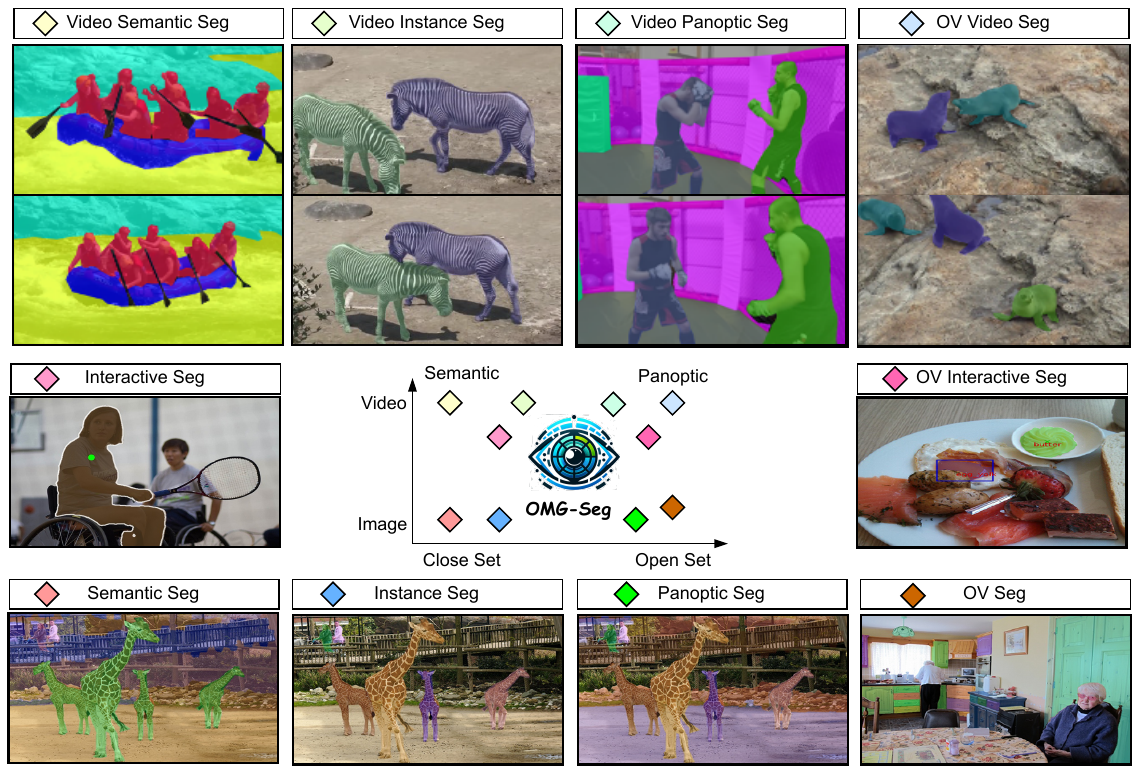}
\vspace{-3mm}
\captionof{figure}{OMG-Seg can handle over ten different segmentation tasks in one framework, including image-level and video-level segmentation tasks, interactive segmentation, and open-vocabulary segmentation. To our knowledge, this is the first model to unify these four directions.}
\label{fig:teaser}
\vspace{0.7cm}
}]

\input{latex/0_abstract}
\input{latex/1_introduction}
\input{latex/2_related_work}

\input{latex/3_method}
\input{latex/4_exp}
\input{latex/5_conclusion}

\input{latex/6_appendix}

{
    \small
    \bibliographystyle{ieee_fullname}
    \bibliography{egbib}
}

\end{document}

%% file: latex/0_abstract.tex
\begin{abstract}

\vspace{-2mm}
In this work, we address various segmentation tasks, each traditionally tackled by distinct or partially unified models. We propose OMG-Seg, One Model that is Good enough to efficiently and effectively handle all the segmentation tasks, including image semantic, instance, and panoptic segmentation, as well as their video counterparts, open vocabulary settings, prompt-driven, interactive segmentation like SAM, and video object segmentation. 
To our knowledge, this is the first model to handle all these tasks in one model and achieve satisfactory performance.
We show that OMG-Seg, a transformer-based encoder-decoder architecture with task-specific queries and outputs, can support over ten distinct segmentation tasks and yet significantly reduce computational and parameter overhead across various tasks and datasets. 
We rigorously evaluate the inter-task influences and correlations during co-training. Code and models are available at \url{https://github.com/lxtGH/OMG-Seg}.

\end{abstract}

%% file: latex/1_introduction.tex
\section{Introduction}
\label{sec:intro}



Visual segmentation that aims to understand semantics at the pixel level has been a longstanding problem~\cite{malik2001contour,schroff2008object} in the vision community, fueling advancements in diverse applications such as robotics, autonomous vehicles, and augmented / virtual reality systems. 
Over the past decade, owing to the tremendous progress in deep learning~\cite{minaee2021image,zhou2023survey,li2023transformer,sfnet,deeplabv3,deeplabv3plus,kirillov2019panopticfpn,li2020improving}, this fundamental problem has been significantly transformed into a diverse set of tasks for image and video data, including basic semantic object / instance segmentation, panoptic segmentation, and the more recent prompt-driven interactive segmentation~\cite{kirillov2023segment}. Consequently, a plethora of task-specific deep segmentation models (\eg, Mask-RCNN~\cite{maskrcnn}, Mask2Former~\cite{cheng2021mask2former}, and SAM~\cite{kirillov2023segment}), along with different benchmarks, have been proposed. The latest studies~\cite{zareian2021open,gu2021open,wu2023open,ghiasi2022scaling} strive to extend these standard close-set segmentation models to more dynamic, real-world scenarios. This involves integrating pre-trained vision-language foundation models, \eg, CLIP~\cite{radford2021learning}, into deep segmentation frameworks, enabling visual segmentation through open-vocabulary text descriptions. 

Most existing deep segmentation models often focus on a single specific task. In many scenarios, a generalizable model capable of handling a broader spectrum of segmentation tasks is highly desired. A unified model would eliminate the necessity for task-specific designs while providing a versatile solution to a wide range of segmentation tasks through a single and cohesive architecture. This approach benefits significantly from leveraging large and varied data corpora, which enhances the model's adaptability and effectiveness across different segmentation tasks.

Unifying diverse segmentation tasks within a single model is non-trivial because each task typically comes with its own unique model design. 
The emergence of transformers~\cite{detr,VIT,liu2022convnet} has catalyzed several segmentation models based on the Detection Transformer (DETR) architecture~\cite{cheng2021mask2former,li2022videoknet,li2023tube,kim2022tubeformer,zhang2021knet,jain2023oneformer}, yielding notable successes in performance and task integration. 
Concurrently, there are also models~\cite{yu2023fcclip,xu2023odise,athar2023tarvis,zou2022xdecoder,UNINEXT, wang2023hierarchical,gu2023dataseg} that employ a similar framework to merge open-vocabulary and multi-dataset segmentation within a unified architecture. 
Yet, these models often fall short in generalizing to video or interactive segmentation, both essential for broader applications. 
Some recent studies~\cite{Painter,SegGPT,wang2023visionllm} aim to unify all vision tasks under one single framework with segmentation included. However, these more generalized models still lag behind task-specific segmentation models in terms of performance.

In this study, we demonstrate that \textit{one model is good enough for all segmentation}\footnote{This includes primarily pure visual and 2D segmentation tasks, excluding specific tasks like medical segmentation~\cite{chen2021transunet} and referring segmentation~\cite{mao2016generation}. Nonetheless, these could be seamlessly integrated into our OMG-Seg framework with appropriate input adaptations.} by introducing OMG-Seg, a unified segmentation model designed to deliver competitive performance across a broad spectrum of visual segmentation tasks.
Unlike previous unified models that typically employ a shared visual backbone but several task-specific branches, OMG-Seg adopts a shared encoder-decoder architecture.
In particular, we unify all the task outputs as a unified query representation. One query can represent a mask label, an image or tube mask, a unique ID, and a visual prompt. 
Then, we can adopt a shared decoder to process all types of queries with their features.
This setup facilitates general training and inference processes that unify all visual-only segmentation tasks, capitalizing on the extensive parameter sharing across tasks. Through co-training on combined image and video datasets, OMG-Seg, once trained, is capable of handling up to ten diverse segmentation tasks across different datasets.

OMG-Seg achieves comparable results on image, video, open-vocabulary, and interactive segmentation settings over eight different datasets, including COCO~\cite{coco_dataset}, ADE-20k~\cite{ADE20K}, VIPSeg~\cite{miao2021vspw}, Youtube-VIS-2019~\cite{vis_dataset}, Youtube-VIS-2021, and DAVIS-17~\cite{caelles20182018}, based on one single shared model. To the best of our knowledge, we are the first to achieve four different settings in one single model.

%% file: latex/2_related_work.tex
\section{Related Work}
\label{sec:relatedwork}





\begin{table*}[t!]
   \centering
    \caption{Setting Comparison For Different Models. We include several representative methods here. Our proposed OMG-Seg can perform various segmentation tasks in one model.}
   \scalebox{0.60}{
   \setlength{\tabcolsep}{3.6mm}{\begin{tabular}{c c c  c c c c c c c c c}
      \toprule[0.15em]
        Methods & SS & IS & PS & VSS & VIS & VPS &  VOS & Open-Set & Multi dataset training & Interactive  & Shared model \\
        \hline
         DeeplabV3+~\cite{deeplabv3plus} & \checkmark &   \\
         MaskRCNN~\cite{maskrcnn} & & \checkmark & \\
        PanopticFPN~\cite{kirillov2019panopticfpn}  & & & \checkmark \\
         DERT~\cite{detr}  & & & \checkmark  \\
        DetectorRS~\cite{qiao2021detectors} & & \checkmark & \checkmark &  \\
        \hline
        TCB~\cite{miao2021vspw}  & & & & \checkmark \\
        VisTR~\cite{VIS_TR} & & & & & \checkmark \\
        VPSNet~\cite{kim2020vps} &  & & & & & \checkmark \\
        STM~\cite{oh2019video_stm_vos} &  & & & & & & \checkmark \\
        \hline
         K-Net~\cite{zhang2021knet} & \checkmark & \checkmark & \checkmark\\
         Mask2Former~\cite{cheng2021mask2former} & \checkmark & \checkmark & \checkmark\\
         Video K-Net~\cite{li2022videoknet} & & & & \checkmark & \checkmark & \checkmark \\
         Tube-Link~\cite{li2023tube}  & & & & \checkmark & \checkmark & \checkmark \\
         TubeFormer~\cite{kim2022tubeformer}  & & & & \checkmark & \checkmark & \checkmark \\
         OneFormer~\cite{jain2023oneformer} & \checkmark & \checkmark & \checkmark &  &&&&&&&  \checkmark \\ 
         TarViS~\cite{athar2023tarvis} & & & & \checkmark & \checkmark & \checkmark & \checkmark &  &  &  &  \checkmark   \\
         \hline
         MSeg~\cite{lambert2020mseg} & \checkmark & & & & & & & & \checkmark & & \checkmark \\
         UNINEXT~\cite{UNINEXT} & & \checkmark & & & \checkmark & & \checkmark & & \checkmark \\
         OpenSeg~\cite{ghiasi2022scaling} & \checkmark & & & & & &  & \checkmark & \checkmark & & \checkmark \\
         SAM~\cite{kirillov2023segment} & & & & & &  & & \checkmark &  & \checkmark \\
         Semantic-SAM~\cite{li2023semanticsam} & \checkmark &\checkmark &\checkmark & & &  & & \checkmark & \checkmark & \checkmark & \checkmark \\
         SEEM~\cite{zou2023segment} &  \checkmark &\checkmark &\checkmark & & & & & \checkmark & \checkmark & \checkmark & \checkmark \\
         OPSNet~\cite{chen2023opsnet} & & & \checkmark & & & & & \checkmark \\
         FreeSeg~\cite{qin2023freeseg} & \checkmark &\checkmark &\checkmark & & & & & \checkmark & & & \checkmark \\
         \hline
        \textbf{OMG-Seg} & \checkmark & \checkmark & \checkmark & \checkmark & \checkmark & \checkmark & \checkmark & \checkmark  & \checkmark & \checkmark & \checkmark \\
      \bottomrule[0.10em]
   \end{tabular}}}
   \label{tab:model_scope}
\end{table*}

\noindent
\textbf{Universal Image/Video Segmentation.} 
The advent of vision transformers~\cite{detr,VIT,liu2022convnet} has led to a wave of innovation in universal segmentation. Recent works~\cite{zhang2021knet,wang2020maxDeeplab,cheng2021mask2former,yu2022kmaxdeeplab,panopticpartformer,cheng2023segment,li2022deep,guo2022beyond,guo2022segnext,yang2021collaborative,yang2024aost} have developed mask classification architectures grounded in an end-to-end set prediction approach, outperforming specialized models~\cite{htc,kirillov2019panoptic,maskrcnn,sfnet,VLT,Li2022SFNetFA,MOSE,MeViS,hu2023suppressing} in both image and video segmentation tasks~\cite{kim2022tubeformer,li2022videoknet,li2023tube}. Despite these advancements, most existing methods still rely on distinct models for different segmentation tasks and datasets.
Recently, there has been a shift towards training a single model~\cite{jain2023oneformer,xu2024rapsam,UNINEXT,gu2023dataseg} across diverse datasets and tasks, reaping the benefits of parameter sharing. For instance, OneFormer~\cite{jain2023oneformer} integrates three image segmentation tasks within a single model, while UNINEXT~\cite{UNINEXT} concentrates on unifying instance-level tasks. Similarly, TarVIS~\cite{athar2023tarvis} combines various video segmentation tasks using target prompts. However, none of these existing works has thoroughly investigated the joint training of image, video, and prompt-driven data within one comprehensive segmentation model. Our work stands as the first attempt in this direction, stretching the potential of co-training across these domains. For a more in-depth comparison of model capabilities, please refer to Tab.~\ref{tab:model_scope}.

\noindent
\textbf{Visual Foundation Models.} Recent studies in visual foundation models have exhibited a diversification in optimization techniques, encompassing various learning paradigms. These include vision-only pre-training strategies \cite{MaskedAutoencoders2021,li2023correlational,xie2023masked}, joint vision-language pre-training approaches\cite{EVA,li2022blip}, and multi-modal frameworks that incorporate visual prompting~\cite{SegGPT,alayrac2022flamingo,peng2023kosmos}. A notable example, SAM~\cite{kirillov2023segment}, demonstrates the generalizability and scalability of extensive training in achieving general segmentation. Building on this, Semantic-SAM~\cite{li2023semanticsam} augments the SAM model by adding semantic labels and increased levels of granularity. However, despite their impressive capabilities, these visual foundation models typically fall short in video segmentation tasks, necessitating further refinement for optimal performance in more dynamic contexts.

\noindent
\textbf{Open Vocabulary Segmentation.} This line of visual segmentation research~\cite{LSeg,wang2021unidentified} aims to recognize and segment novel objects beyond the limited closed-set visual concepts. Leveraging the transferrable representations offered by vision language models (VLMs), many studies~\cite{zareian2021open,gu2021open,wu2023open,zang2022open,ghiasi2022scaling,wang2023detecting,wu2023betrayed,yuan2024ovsam,xu2023dst,wu2023clipself,zhou2023rethinking} explore the alignment between region and text representations during training. At the inference stage, detectors can recognize new classes using the text embeddings derived from VLMs. Our model follows this notion to achieve open vocabulary segmentation. In particular, we use frozen VLMs to serve both as a feature extractor and classifier. This strategy allows for a seamless transition into the open vocabulary setting.

\noindent
\textbf{Unified Modeling.} The adaptable nature of the transformer architecture~\cite{vaswani2017attention,VIT} facilitates the sharing of fundamental modules across various modalities. This versatility has inspired several research initiatives that use a common transformer framework for different domains. Notably, efforts in the realm of vision generalists have been directed toward unifying disparate tasks within the vision domain. For instance, the Pix2Seq series~\cite{chen2022unified, chen2021pix2seq} approach task unification through auto-regressive token prediction. Similarly, Unified-IO~\cite{lu2022unified} implements a sequence-to-sequence pipeline, converting diverse inputs and outputs into discrete token sequences. Furthermore, recent advancements~\cite{Painter,SegGPT,bar2022visual,fang2023explore,bai2023sequential,wang2023skeleton} have explored visual in-context learning as the means to combine various vision tasks.
These methods predominantly target task unification across domains. However, bridging the performance gap between unified segmentation models and purpose-built segmentation models remains an open problem.

%% file: latex/3_method.tex
\section{Methodology}
\label{sec:method}

\begin{figure*}
    \centering
    \includegraphics[width=0.80\textwidth]{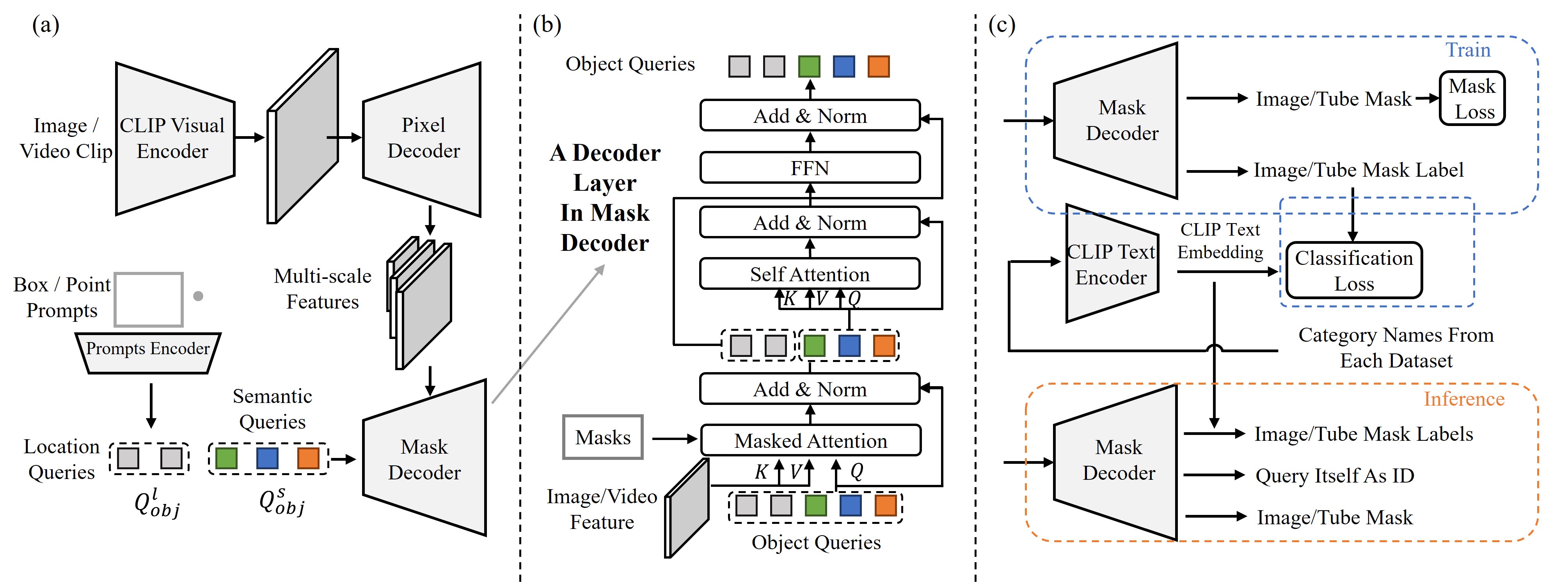}
    \caption{OMG-Seg meta-architecture. \textbf{(a)} OMG-Seg follows the architecture of Mask2Former~\cite{cheng2021mask2former}, containing a backbone (CLIP Visual Encoder), a pixel decoder, and a mask decoder. The different parts are a shared mask decoder for both image and video segmentation and a visual prompt encoder. We use two types of mask queries, i.e., semantic queries, for instance/semantic masks or mask tubes, and location queries that encode box or point prompts. \textbf{(b)} One decoder layer in the Mask Decoder. The location queries skip the self-attention operation as they are only conditioned on the image content and the location prompts. \textbf{(c)} The forward pass of OMG-Seg in training and inference. We use CLIP's text encoder to represent category names and classify masks by calculating cosine similarity between mask features and text embeddings.}
    \label{fig:method}
    \vspace{-3mm}
\end{figure*}

\noindent
\textbf{Motivation and Overview.} Our OMG-Seg is a single yet versatile model---with reduced task-specific customization and maximal parameter sharing---that can support a diverse set of segmentation tasks, making it one model for all segmentation. Our goal is not to pursue state-of-the-art results for each task but to increase the modeling capacity of one generalizable segmentation model while allowing extensive knowledge sharing between tasks. 

The main idea of our approach is to leverage object queries for representing distinct entities, encompassing various mask types and their respective video formats. 
In Sec.~\ref{sec:method_unified_task}, we begin by reexamining the definitions of image, video, interactive, and open vocabulary segmentation settings. In this exploration, we show that the target outputs of these varied settings can be effectively transformed into a unified query representation. Specifically, a single query can encapsulate a mask label, an image or tube mask, a unique identifier, or a visual prompt.

For example, video segmentation tasks require only an additional ID compared to image segmentation, which can be adapted from the query ID. This allows us to employ a shared decoder to process each query and its associated features in a streamlined manner, with the primary distinction being the specific feature inputs used in cross-attention layers.
In the context of image tasks, we follow the established design of Mask2former~\cite{cheng2021mask2former}, enabling queries and features to engage in masked-cross attention. For video tasks, we incorporate temporal features with 3D position embeddings and focus on predicting tube masks for objects across short video clips. For interactive segmentation tasks, we employ the same decoder as image tasks but skip the self-attention operation to condition mask prediction only on the visual prompts and image contents, as detailed in Sec.~\ref{sec:method_unified_archi}.

In addition, to circumvent class taxonomy conflicts, we adopt CLIP embeddings for mask classification. We employ the frozen CLIP visual encoder as the backbone, whose features are shared by the pixel decoder and the open-vocabulary mask classification. This design enables efficient open-vocabulary inference without incurring additional costs. The training and inference pipelines built on such a frozen backbone are described in Sec.~\ref{sec:method_train_inference}.

\subsection{Unified Task Representation}
\label{sec:method_unified_task}


\noindent
\textbf{Image Segmentation.} Given an input image $ I \in \mathbb{R}^{H\times {W}\times 3}$, the goal of image segmentation is to output a group of masks $\{y_i\}_{i=1}^G = \{(m_i, c_i)\}_{i=1}^G \,$ where $c_i$ denotes the class label of the binary mask $m_i$ and $G$ is the number of masks, ${H}\times {W}$ are the spatial size. According to the scope of class labels and masks, we report the results of three different segmentation tasks, including semantic segmentation (SS), instance segmentation (IS), and panoptic segmentation (PS). PS is the unification of both SS and IS, which contains countable thing classes and uncountable stuff classes. For all three tasks, we adopt mask classification architecture~\cite{cheng2021maskformer,zhang2021knet}, where each mask corresponds to a semantic label.


\noindent
\textbf{Video Segmentation.} Given a video clip input as $ V \in \mathbb{R}^{T\times H\times {W}\times 3}$, where $T$ represents the frame number, the goal of video segmentation is to obtain a mask tube $\{y_i\}_{i=1}^N = \{(m_i, c_i, d_i)\}_{i=1}^N \,$, where $N$ is the number of the tube masks $m_i \in {\{0,1\}}^{{T}\times {H}\times {W}}$. $c_i$ denotes the class label of the tube mask $m_i$ while $d_i$ denotes the instance ID of each tube mask. Each tube mask can be classified into a countable thing class or uncountable stuff class, where the thing classes are also assigned a unique ID. For stuff masks, the tracking is zero by default. When $N=C$ and the task only contains stuff classes, and all thing classes have no IDs, VPS turns into video semantic segmentation (VSS). If ${\{y_i\}_{i=1}^N}$ overlap and $C$ only contains the thing classes and all stuff classes are ignored, VPS turns into video instance segmentation (VIS). Video Object Segmentation (VOS) aims to track the first framework masks without performing classification. Motivated by image segmentation, we also adopt the tube mask classification architecture~\cite{li2023tube, kim2022tubeformer} to train and link short tubes along the temporal dimension. For VOS, we adopt class-agnostic tube-wised training, which is similar to VPS and VIS. 

\noindent
\textbf{Interactive Segmentation.} The interactive segmentation in SAM~\cite{kirillov2023segment} framework takes both image $I$ and visual prompts $P \in \mathbb{R}^{N \times \{2, 4\}}$, such as points and boxes, as inputs, and it outputs the corresponding binary image masks $\{y_i\}_{i=1}^N = \{m_i \in H\times {W} \}_{i=1}^N \,$ $N$ is the number of visual prompts. Each visual prompt is encoded into an object query, which naturally can be the input of the decoder, like in ~\cite{kirillov2023segment, cheng2021mask2former}. In our experiments, we use the shared decoder for all different task queries.

\noindent
\textbf{Open-Vocabulary and Multi-Dataset Segmentation.} The task formulation is the same as the previous image and video segmentation. However, this setting goes beyond fixed label space. In particular, it requires open-set recognition on various datasets. Meanwhile, multi-dataset segmentation requires one model to segment more concepts under different datasets. As a common practice, we adopt CLIP text embedding as the mask classifier, which avoids taxonomy conflicts and achieves open-set recognition at the same time. As a result, we measure the distance between the visual query feature and class embeddings rather than the learned classifier.

\noindent
\textbf{All the Things are in Queries.} As mentioned above, by combining all different settings, we can represent all the output segmentation entities using the same query-based mask classification framework. In particular, one object query corresponds to one mask $m_{i}$, label $c_{i}$, and ID $d_{i}$. Depending on different task settings, the formats and ranges of $m_{i}$, $c_{i}$, and $d_{i}$ are different. However, the formats and ranges of $m_{i}$, $c_{i}$, and $d_{i}$ are similar. Thus, it is natural to put all these tasks into one shared encoder-and-decoder framework and co-train one model for all segmentation tasks
Thus, it is natural to put all these tasks into one shared encoder-and-decoder framework and co-train one model for all segmentation tasks. 


\subsection{OMG-Seg Architecture}
\label{sec:method_unified_archi}


\noindent
\textbf{Overview.} OMG-Seg follows the architecture design of Mask2Former~\cite{cheng2021mask2former}. As shown in Fig.~\ref{fig:method}, it contains a backbone, a pixel decoder, and a mask decoder. 
The difference lies in the following aspects, including frozen backbone design, combined object queries which contain both object query and visual prompt, and a shared multi-task decoder. 
Given different task settings, the decoder outputs corresponding masks and labels. We 


\noindent
\textbf{VLM Encoder as Frozen Backbone.} To enable open-vocabulary recognition, for the backbone part, we adopt the frozen CLIP visual model as a feature extractor. We use the ConvNeXt architecture~\cite{liu2022convnet} from the OpenCLIP~\cite{openclip_paper}. Given image/video inputs, the VLM encoder extracts multi-scale frozen feature $\{F^{frozen}_{j} \}^{3}_{j=1}, $ for further process.

\noindent
\textbf{Pixel Decoder as Feature Adapter.} The pixel decoder is the same as Mask2Former, which contains multi-stage deformable attention layers. It transforms the frozen feature $\{F^{frozen}_{j} \}^{3}_{j=1}, $ into the fused feature $\{F^{fuse}_{j} \}^{3}_{j=1}, $ with the same channel dimension, where $j$ is the layer index of feature. $j=3$ is the highest-resolution feature.

\noindent
\textbf{Combined Object Queries.} As analyzed above, each object query represents one type of mask output. However, from the functionality perspective, image, video, and interactive modes represent different properties. For images, object queries focus on object-level localization and recognition. For video, object queries may involve temporal consistency, such as the same object long different frames.
For interactive segmentation, object queries are forced to locate specific regions. For image and video input, we adopt object queries to represent image masks or tracked tube masks. Since both need semantic labels. We term them as semantic queries, $Q_{obj}^{s}$. For interactive mode, following SAM~\cite{kirillov2023segment}, we adopt the prompt encoder to encode the various visual prompts into the same shape of object queries. We term them as location queries, $Q_{obj}^{l}$. Thus, we can share the same interface for the transformer decoder.

\noindent
\textbf{Shared Multi-Task Decoder.} Its main operation is cross-attention, which takes in the combined object queries ($Q_{obj}^{s}$ and $Q_{obj}^{l}$) and the image/video feature $F^{fuse}_{j}$, and outputs refined object queries. The final masks are obtained via dot-product of refined queries and high-resolution feature $F^{fuse}_{3}$.
For image semantic level tasks, we adopt the same procedure of Mask2Former. In particular, $Q_{obj}^{s}$ perform masked cross-attention~\cite{cheng2021mask2former} with multi-scale features $F^{fuse}_{j}$. $Q_{obj}^{s}$ is Query while $F^{fuse}_{j}$ are the Key and Value. Then, a multi-head self-attention (MHSA) layer is applied to the refined queries. The refined queries and high-resolution features are used to 

For video tasks, we adopt the same cross-attention design. The only difference is the pyramid features $F^{fuse}_{j}$ are contacted along the temporal dimension with 3D position embeddings, which are the default setting as previous works~\cite{cheng2021mask2former_vis,li2023tube}. The combined video features and refined queries are used to predict the tube mask. 

For interactive segmentation, we carry out the same cross-attention design. However, we skip the self-attention to avoid interaction between mask queries in the MHSA layer, since the interactive segmentation only cares about the input visual prompt regions. After obtaining the refined object query, it is passed through a prediction FFN, which typically consists of a 3-layer perceptron with a ReLU activation layer and a linear projection layer. All the queries are supervised by mask classification loss and mask prediction loss. The decoding process is in a cascaded manner, in three stages for each feature pyramid.


\subsection{Training and Inference}
\label{sec:method_train_inference}

\noindent
\textbf{Joint Image Video Dataset Co-training.} Rather than first pre-trained on image datasets, our goal is to train all segmentation tasks only once jointly. All training targets are one entity label and mask for all three different cases. The entity can be thing, stuff, class-agnostic masks, and their corresponding labels. Note that the instance masks with the same ID $d$ form the tube masks. During training, we apply Hungarian matching between the predicted and ground-truth entity masks to assign object queries to video/image entities, and then supervise their predicted masks and classification. The classifier is replaced by CLIP text embedding to avoid cross-dataset taxonomy conflicts. The final loss function is given as $L = \lambda_{cls}L_{cls} + \lambda_{ce}L_{ce} + \lambda_{dice}L_{dice}$. Here, $L_{cls}$ is the Cross-Entropy (CE) loss for mask classification, and $L_{ce}$ and $L_{dice}$ are mask Cross Entropy (CE) loss and Dice loss~\cite{dice_loss,wang2019solo} for segmentation, respectively. 

\noindent
\textbf{Universal Inference.} For image segmentation, we follow the same inference procedure of Mask2Former~\cite{cheng2021mask2former}. For example, for PS, we merge the things and stuff according to the sorted scores. The scores are generated by CLIP text embedding. For video segmentation tasks, for VIS and VPS, to generate instance ID, following previous work, we use query matching rather than introducing extra tracking components. For VOS tasks, we adopt mask matching between the first frame and the remaining frames. For interactive segmentation tasks, we follow the original SAM~\cite{kirillov2023segment}, by providing box and point prompts, and obtain the binary masks. For open vocabulary segmentation, since we have a frozen CLIP encoder, we merge mask pooled score and learned score with the open-vocabulary embeddings.

\noindent
\textbf{Combining Tasks For More Applications.} Since our model can perform various segmentation tasks, combining interactive, open vocabulary and image/video segmentation tasks can lead to several new applications. For example, we can combine interactive and video segmentation, leading to flexible prompt-driven video object segmentation. Or we can combine interactive segmentation with an open vocabulary setting, which results in open vocabulary interactive segmentation. More examples are provided in Sec.~\ref{sec:exp} and supplementary.

%% file: latex/4_exp.tex
\section{Experiments}
\label{sec:exp}

\begin{table*}[t]
   \centering
    \caption{Experiment results of OMG-Seg on image, video, open-vocabulary, and SAM-like settings. * denotes models are pre-trained on the Object365 dataset~\cite{shao2019objects365}. We only list representative methods due to the page limit. Refer to the supplementary material for more methods. Our results are the \textbf{averaged results} of five different experiments, due to the dataset noises.}
   \scalebox{0.55}{
   \setlength{\tabcolsep}{1.7mm}{\begin{tabular}{c c c  c c c c c c c c c c c c c c }
      \toprule[0.15em]
         \multirow{2}{*}{Methods} &  \multirow{2}{*}{Backbone} & COCO-PS & Cityscapes-PS & COCO-IS & VIPSeg-VPS & YT-VIS-19 & YT-VIS-21-OV & ADE-OV  & DAVIS-17-VOS-OV & COCO-SAM & Share Model \\
         &   & PQ & PQ & mAP & VPQ & mAP & mAP & PQ & J\&F & mIoU & - \\
        \hline
         DetectorRS~\cite{qiao2021detectors} & ResNet50 & - & - &42.1 &-&-&-&-&-&-&- \\
         HTC~\cite{htc} & ResNet50  &- & - &  38.4  &-&-&-&-&-&- &- \\
         STM~\cite{stm_vos} & ResNet101 & - & - &-&-&-&-& - &  79.2 &-&- \\
         \hline
         K-Net~\cite{zhang2021knet} & ResNet50 & 47.1 & -  & 38.6 &- &-&-&-&-&-&- \\
         Mask2Former~\cite{cheng2021mask2former} & ResNet50 & 51.9 & 62.1 & 43.7 &-&-&-&-&-&-&- \\
         Mask2Former~\cite{cheng2021mask2former} & Swin-Large & 57.8 & 66.6 & 50.1 &-&-&-&-&-&-&- \\
         k-Max Deeplab~\cite{yu2022kmaxdeeplab} & ResNet50 & 53.0 & 64.3 &-&-&-&-&-&-&- &-\\
         k-Max Deeplab~\cite{yu2022kmaxdeeplab} & ConvNeXt-Large & 58.1 & 68.4  &-&-&-&-&-&-&-&-\\
         \hline
         SeqFormer~\cite{seqformer} & ResNet50 & - & -& -& -& 47.4 & - &-& - &-&-\\
         IDOL~\cite{IDOL} & Swin-Large & - & -&- & -& 64.3 & - &-&-&-&-\\
         MinVIS~\cite{huang2022minvis} & Swin-Large  & - & -& -& -& 61.6 & -&-&-&- &- \\
         Video K-Net~\cite{li2022videoknet} & ResNet50 & - &  -& -& 26.1& 40.5&-&-&-&-&-  \\
         Tube-Link~\cite{li2023tube} & ResNet50 & - & -& -& 41.2 & 52.8 & -&-&-&- &- \\
         Tube-Link~\cite{li2023tube} & Swin-base & - &-&-&54.5&-&-&-&-&- &- \\
         OneFormer~\cite{jain2023oneformer} & Swin-Large & 58.0 & 67.2 & 49.2 &- &- &- &- & -&- & \checkmark \\ 
         TarViS~\cite{athar2023tarvis} &  Swin-Large & - & - & - & 48.0 & -&- &- &-  & -& \checkmark\\
         \hline
         fc-clip~\cite{yu2023fcclip}  & ConvNeXt-Large & 54.4 & - & 44.6  & -& -& -& 26.8 &- & -&  \checkmark  \\
         ODISE~\cite{xu2023odise} &  ViT-Large & 55.4 & - & 46.0 &- & -& -&  22.6 &- &- & \checkmark \\
         DaTaSeg~\cite{gu2023dataseg} & ViT-L & 53.5 & - & -&-&-&-&-&-&-& \checkmark \\
         \hline
         X-Decoder~\cite{zou2022xdecoder} & DaViT & 56.9 & - & 46.7 & - & -& -& 21.8 & -& -& \checkmark\\
         SEEM~\cite{zou2023segment} * & DaViT & 57.5 & -  & 47.7 & -&- &- &- & 58.9 &  83.4 & \checkmark \\
         UNINEXT~\cite{UNINEXT} *  & ConvNeXt-L  &- & - & 49.6 & -&  64.3 &-&-& 77.2 &-& \checkmark\\
         HIPIE~\cite{wang2023hierarchical} * & ViT-H & 58.0 & - & 51.9  &- &- &- & 20.6&-&-& \checkmark \\
         OpenSeeD~\cite{zhang2023simple} * & Swin-L  & 59.5 & -  & 53.2 & - & - & - & 19.7  & - & - & \checkmark \\ 
         SAM~\cite{kirillov2023segment} & ViT-H & - & - & - & - & - & - & - & - & 55.3 & \checkmark \\
         Semantic-SAM~\cite{li2023semanticsam} & Swin-T & 55.2 & - & 47.4 & - & - & - & - & - & 53.0 & \checkmark  \\
         Painter~\cite{Painter} & ViT-L & 43.4 & - & - & - & - & - & - & - & - & \checkmark\\
         \hline
        OMG-Seg & ConvNeXt-Large (frozen) & 53.8 & 65.7 & 44.5 & 49.8 & 56.4  & 50.5  & 27.9 & 74.3 & 58.0 & \checkmark   \\
        OMG-Seg & ConvNeXt-XX-Large (frozen) &  55.4 & 65.3  & 46.5 &  53.1  & 60.3  &  55.2  & 27.8  & 76.9 & 59.3 &  \checkmark  \\
      \bottomrule[0.10em]
   \end{tabular}}}
   \label{tab:main_result_tab}
\end{table*}
\noindent
\textbf{Datasets and Metrics.} Unlike regular settings, we aim to explore co-training on multiple datasets as much as possible. 
In Tab.~\ref{tab:main_result_tab}, we use COCO panoptic~\cite{coco_dataset}, COCO-SAM, VIPSeg~\cite{miao2022large}, and Youtube-VIS-2019~\cite{vis_dataset} (YT-VIS-19) as training datasets. In addition to the closed-set testing, 
we include the open vocabulary (OV) inference by using Youtube-VIS-2021, ADE-20k~\cite{ADE20K}, and DAVIS-2017 datasets~\cite{caelles20182018}, where their annotations are not used during the training.
COCO-SAM is created by using the ground truth boxes, and mask center points are visual prompts. The annotations are obtained by COCO panoptic masks. Moreover, we also include the multi-dataset settings in Tab.~\ref{tab:main_result_tab_multi_dataset} to verify the effectiveness of multi-dataset co-training of our OMG-Seg. In addition to Tab.~\ref{tab:main_result_tab}, we add more datasets, including ADE-20k and YT-VIS21 for joint co-training. We use the corresponding metrics for each dataset, including PQ~\cite{kirillov2019panoptic}, mask mAP~\cite{coco_dataset}, VPQ~\cite{kim2020vps}, tube mAP~\cite{vis_dataset}, J\&F~\cite{caelles20182018}, and mIoU~\cite{ADE20K}. 


\noindent
\textbf{Implementation Details.} We implement our models and all other baselines in MMDetection~\cite{chen2019mmdetection}. We use the distributed training framework with 32 A100 GPUs. Each mini-batch has one image per GPU. For data augmentation, we adopt large-scale jitter as previous works~\cite{cheng2021mask2former,li2023tube} to build strong baselines. For all models in each table, we adopt the same training steps. We use OpenCLIP~\cite{radford2021learning} to initialize the backbone network and replace learned classifiers with their corresponding text embeddings. For image inputs, we treat them as pseudo videos by concatenating two images and their masks into one. We adopt different sampling rates to balance the training examples for each dataset. We report results of both frozen and trained backbones for reference. We list more details in the supplementary material.

\begin{table*}[t]
   \centering
    \caption{Experiment results of OMG-Seg on multiple dataset settings. We use five different datasets for balanced joint co-training for only 12 epochs. We also implement compared baselines in the same codebase. }
   \scalebox{0.70}{
   \setlength{\tabcolsep}{2.5mm}{\begin{tabular}{c c c  c c c c c c c c c c c c c c }
      \toprule[0.15em]
        Methods /\ Settings & Backbone & COCO-PS & COCO-IS & ADE-PS & VIPSeg-VPS & YT-VIS-19 & YT-VIS-21 & Params(M) &  Share Model \\
        \hline
         K-Net~\cite{zhang2021knet} & ConvNeXt-Large (trained) & 50.5  &  42.3 &  40.2 &  - &  - &  -  &   & -    \\
         Mask2Former~\cite{cheng2021mask2former} & ConvNeXt-Large (trained) & 53.2 &  45.2 & 43.2  & - & - &  -  &  & - \\
         Mask2Former-VIS~\cite{cheng2021mask2former_vis} & ConvNeXt-Large (trained) & - & - & - &  - & 45.8  & 42.3 &  \\
        \hline
        single dataset baseline & ConvNeXt-Large (frozen) & 52.5 & 45.6 &  41.2 & 42.3 & 45.3 & 44.3  & 1326  & -  \\
        OMG-Seg & ConvNeXt-Large (frozen) & 52.9 &  44.3  &  28.2  &  46.9  &  48.8  &  46.2   &  221 & \checkmark    \\
         OMG-Seg & ConvNeXt-Large (trained) & 55.0  &  45.3   &  36.8  & 45.8   &  47.2  & 45.2  & 221 & \checkmark  \\ 
      \bottomrule[0.10em]
   \end{tabular}}}
   \label{tab:main_result_tab_multi_dataset}
\end{table*}

\subsection{Main Results}


\noindent
\textbf{System-level Comparison.} 
In Tab.~\ref{tab:main_result_tab}, we present a comparative analysis of our OMG-Seg against recent methodologies across a variety of settings. A significant highlight of our work is its unique capability to deliver substantial results in all scenarios using a \emph{single} model framework.
In the realm of specific image and video segmentation models, OMG-Seg demonstrates performance on par with leading approaches like Mask2Former~\cite{cheng2021mask2former}, Tube-Link~\cite{li2023tube}, and TarViS~\cite{athar2023tarvis}. While it exhibits a slight decrease in performance on the COCO image segmentation benchmark, it achieves near state-of-the-art results on the VIPSeg datasets, showcasing its robustness and versatility.
Furthermore, when benchmarked against open vocabulary methods such as FCCLIP~\cite{yu2023fcclip} and ODISE~\cite{xu2023odise}, OMG-Seg not only competes favorably but also outperforms ODISE in certain scenarios. This is particularly evident in the realm of open vocabulary video segmentation on YT-VIS-21, as detailed in the 7th column of the table. These findings underscore the effectiveness and adaptability of our OMG-Seg approach in handling a wide array of segmentation challenges.

In addition, our method has been benchmarked against recent unified models, revealing insightful comparisons. When compared with vision generalists such as that described in~\cite{Painter}, our approach, OMG-Seg, demonstrates superior performance. However, in comparison with several specialized segmentation models, including UNINEXT~\cite{UNINEXT} and Wang et al.~\cite{wang2023hierarchical}, we observe a discernible performance discrepancy in the COCO datasets, notably in panoptic and instance segmentation tasks.
This gap, we argue, can be partially attributed to our training regime, which spans only 24 epochs, and also we keep the backbone frozen.
%
Furthermore, the integration of video segmentation and interactive segmentation datasets for joint co-training presents a more formidable challenge compared to previous works. This is primarily because learning spatial-temporal and localization-sensitive features from image data is inherently more complex, given the diversity of the learning targets.

Despite these challenges, it is noteworthy that no other existing models offer the comprehensive segmentation capabilities that OMG-Seg does. This ability to effectively handle all forms of segmentation, despite the small performance gaps noted, reinforces our assertion that OMG-Seg is a robust and versatile model suitable for diverse segmentation scenarios.


%
\noindent
\textbf{Multi-dataset Setting.} 
%
In Tab.~\ref{tab:main_result_tab_multi_dataset}, we extend our investigation to multi-dataset settings. To ensure a fair comparison in the same setting, we reimplemented two key baselines: K-Net~\cite{zhang2021knet} and Mask2Former~\cite{cheng2021mask2former}. Our findings indicate that joint co-training generally enhances performance across most video segmentation datasets, leading to substantial model parameter reduction (from 1326M to 221M). This improvement is consistent across three VPS and VIS datasets, irrespective of whether the backbones are frozen or not.
%
However, it is noteworthy that the performance on the ADE-20k dataset significantly diminishes under joint co-training. We hypothesize that this is largely due to the challenges posed by scale variance and the uneven distribution of classes within the dataset. Interestingly, when using a pre-trained backbone, we observe an uplift in image segmentation performance, albeit at the cost of a minor decline in video segmentation efficacy. This trade-off can be attributed to the unbalanced nature of samples that pursue different optimization objectives, essentially causing a tug-of-war over the representational capacity of the backbone. Such a scenario suggests that incorporating a greater volume of video training examples could potentially address this issue.

\begin{figure}[!t]
    \centering
   \includegraphics[width=0.5\textwidth]{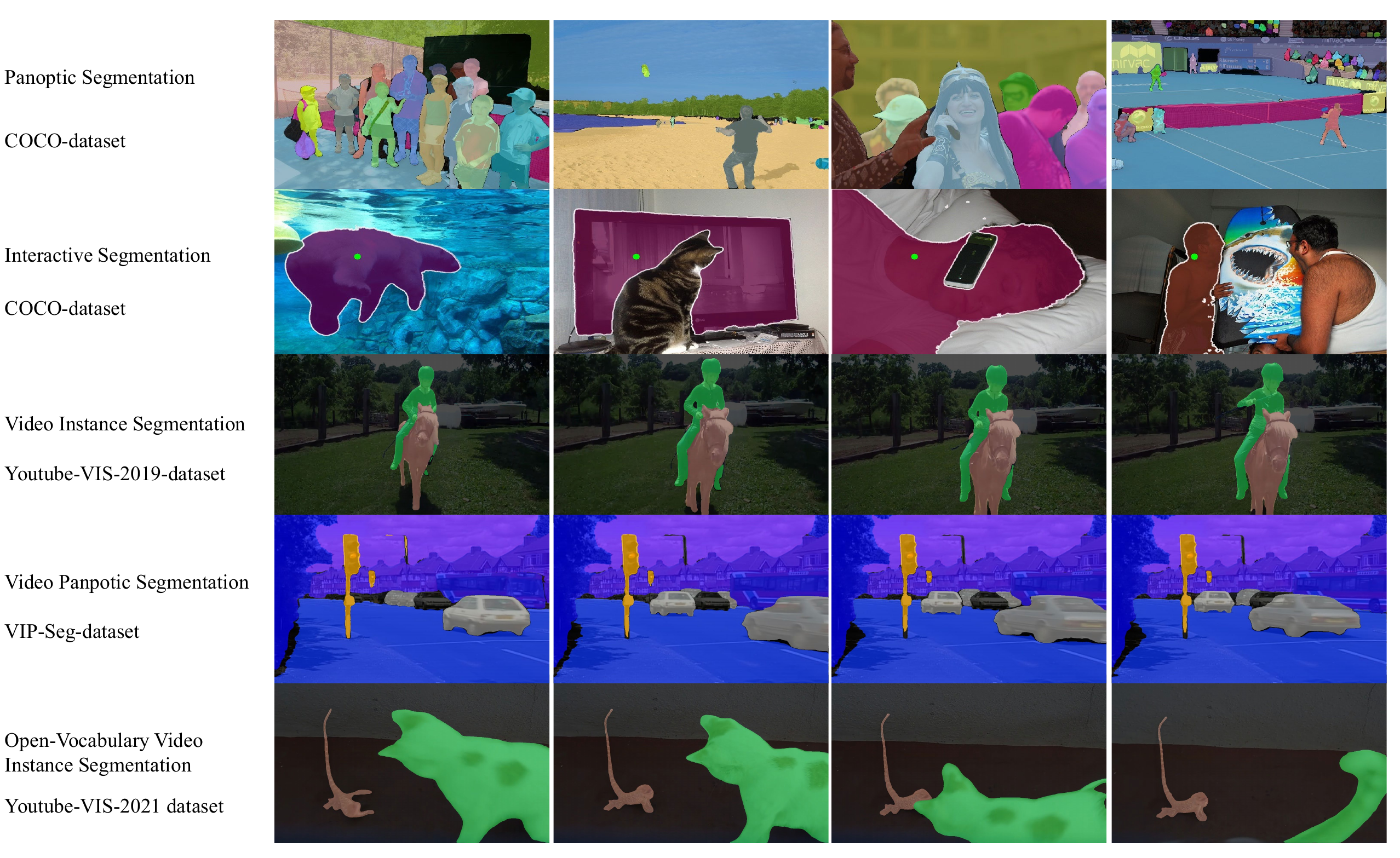}
    \caption{Functional Visualization of OMG-Seg model. We list five different tasks from four datasets as examples. Our method achieves high-quality segmentation, tracking, and as well as interactive segmentation in one shared model.}
    \label{fig:visual_results}
\end{figure}

\noindent
\textbf{Qualitative Result.} 
%
In Fig.~\ref{fig:visual_results}, we show the effectiveness of our OMG-Seg model using a ConvNeXt-Large model across five different tasks. The first two rows demonstrate the model's high-quality image segmentation capabilities on the COCO dataset. In the VIS and VPS tasks, OMG-Seg shows proficiency in segmenting and tracking foreground objects. Notably, in the last row, we show an open-vocabulary video instance segmentation on Youtube-VIS, successfully identifying the ``lizard'' class, which was not included in the training set.

\subsection{Ablation Study and Analysis}
\label{sec:ablation_analysis}

In this section, we use COCO, VIPSeg, and Youtube-VIS-19 for ablation studies of our OMG-Seg. All experiments use frozen ConvNeXt-Large as the backbone and the same data augmentation with 12 epochs training by default. 

\begin{table}[t!]
\centering
\caption{Ablation on joint co-training. (a), COCO-PS. (b), VIPSeg-VPS. (c). YT-VIS-19.}
\resizebox{0.95\linewidth}{!}{
\begin{tabular}{c|ccccccc}
\toprule[0.2em]
Setting & COCO-PS & VIPSeg-VPS & YT-VIS-19 & ADE-OV & YT-VIS-21-OV \\
\midrule
 a  & 53.4 & 32.2 & 34.2 & 25.5 & 30.3  \\
 a + b & 52.9 & 49.0 & 45.2 & 26.2 & 39.6  \\
 a + b + c & 53.0 & 48.5  & 56.8 & 26.1 & 50.3 \\
\bottomrule[0.2em]
\end{tabular}
}
\label{tab-ablation:effect_of_co_training}
\end{table}

\noindent
\textbf{Effect of Training Dataset.} 
%
In Tab.~\ref{tab-ablation:effect_of_co_training}, we evaluate the impact of various datasets on model performance. As indicated in the first row, using only the COCO dataset yields satisfactory zero-shot results across other datasets, largely attributed to the employment of frozen CLIP visual features for zero-shot region feature classification.
Upon integration of the VIPSeg dataset, a slight dip in performance on the COCO dataset is observed. However, this is counterbalanced by significant improvements in both the VIPSeg and Youtube-VIS datasets.
Incorporating all three datasets, COCO, VIPSeg, and Youtube-VIS, results in an optimal performance balance across all datasets, establishing this combination as our preferred and default configuration.

\begin{table}[t!]
\centering
\caption{Ablation on shared decoder design.}
\resizebox{0.95\linewidth}{!}{
\begin{tabular}{c|ccccccc}
\toprule[0.2em]
Setting & COCO-PS & VIPSeg-VPS & Param & GFlops \\
\midrule
 shared  & 53.0 &  48.5 & 221 & 868 \\
 decoupled image/video & 53.6 & 46.2 & 243 & 868 \\
\bottomrule[0.2em]
\end{tabular}
}
\label{tab-ablation:share_head_design}
\end{table}

\noindent
\textbf{Ablation on Shared Decoder Design.} 
%
In Tab.~\ref{tab-ablation:share_head_design}, we explore the efficacy of a shared decoder design. Employing a separate decoder head for video segmentation tasks results in a slight performance decrease. This outcome is influenced by our use of pseudo-video samples during image dataset training. By sharing the decoder, we align the optimization objectives more closely, which particularly benefits the video datasets with short clips~\cite{vis_dataset}.


\begin{table}[t!]
\centering
\caption{Ablation on whether using extra adapters.}
\resizebox{0.95\linewidth}{!}{
\begin{tabular}{c|ccccccc}
\toprule[0.2em]
Setting & epoch & COCO-PS & VIPSeg-VPS & Params (M) & GFlops (G)  \\
\midrule
baseline & 12  & 53.0 &  48.5 & 221 & 868 \\
+ Adapter~\cite{chen2022vitadapter} & 12 &  53.5 &  49.2 & +11 & +103 \\
+ More Pixel Decoder Layer~\cite{zhu2020deformabledetr} &  12 & 53.6  & 49.4 & +21  & +60 \\
\midrule
baseline & 36 &  54.8 & 50.1 & 221 & 868 \\
+ Adapter~\cite{chen2022vitadapter} & 36 & 54.6 & 49.6 & +11 & +103\\
+ More Pixel Decoder Layer~\cite{zhu2020deformabledetr} &  36 & 54.7  &  50.2 & +21  & +60   \\
\bottomrule[0.2em]
\end{tabular}
}
\label{tab-ablation:ablation_on_feature_adapter}
\end{table}

\noindent
\textbf{Ablation on Extra Adapter.} 
%
In Tab.~\ref{tab-ablation:ablation_on_feature_adapter}, we assess the addition of an extra adapter to the frozen CLIP backbone, enhancing the capacity of OMG-Seg. Our experiments reveal that the adapter~\cite{chen2022vitadapter} boosts performance with fewer training epochs, but its effectiveness against the baseline disappears in extended training scenarios. 
In addition, we experiment with increasing the neck capacity by duplicating attention layers in the pixel decoder, observing similar outcomes to the adapter implementation. Consequently, we opt not to incorporate additional adapters, maintaining a cleaner and simpler framework.

\begin{table}[t!]
\centering
\caption{Ablation on different CLIPs.}
\resizebox{0.95\linewidth}{!}{
\begin{tabular}{c|ccccccc}
\toprule[0.2em]
Backbone & epoch & COCO-PS & VIPSeg-VPS & ADE-OV & Params(M)  & Flops(G) \\
\midrule
ResNet50 & 12 & 44.8 & 42.0 & 18.2 & 59.5 & 340 \\
ConvNeXt Large  & 12  & 53.0 & 48.5 &  26.8 & 221 & 868  \\
ConvNeXt XX-Large  & 12  & 54.3 & 53.2 & 27.2 & 820  & 2854 \\
ConvNeXt XX-Large  & 24  & 55.5 & 53.3 & 27.8 & 820 & 2854 \\
ConvNeXt XX-Large  & 36  & 56.0 & 53.0  & 26.7 & 820 & 2854 \\
\bottomrule[0.2em]
\end{tabular}
}
\label{tab-ablation:ablation_on_different_clip}
\end{table}

\noindent
\textbf{Ablation on Other CLIPs.} 
%
In Tab.~\ref{tab-ablation:ablation_on_different_clip}, following the approach of prior open vocabulary research~\cite{yu2023fcclip,FVLM}, we primarily employ convolution-based CLIP models due to their spatial information handling and adaptability to scale variations across different datasets. 
As we scale up the CLIP model size and extend training steps, we observe improvements across all three datasets. Notably, model convergence is achieved at 24 epochs, faster than in previous studies~\cite{cheng2021mask2former}. This accelerated convergence may be attributed to the model's limited capacity, suggesting that larger models could further elevate performance.

%% file: latex/5_conclusion.tex
\section{Conclusion}
\label{sec:conclusion}

In this study, we introduce the first joint co-training framework for image, video, open-vocabulary, and interactive segmentation. Our solution, OMG-Seg, is a novel yet simple framework that uses a unified query representation and a shared decoder for diverse tasks. For the first time, it is possible to train a single segmentation model capable of performing across ten different tasks with competitive performance compared to task-specific models. This approach significantly reduces both the parameter size and the need for specialized engineering in model design for various applications. We envision that our efficient and versatile framework will serve as a robust baseline for multi-task and multi-dataset segmentation.

\noindent \textbf{Acknowledgment.} This study is supported under the RIE2020 Industry Alignment Fund-Industry Collaboration Projects (IAF-ICP) Funding Initiative, as well as cash and in-kind contributions from the industry partner(s). The project is also supported by Singapore MOE AcRF Tier 1 (RG16/21). 






%% file: latex/6_appendix.tex

    



    


\section{Appendix}

\appendix

\noindent
\textbf{Overview.} In this appendix, we first present more method details in Sec.~\ref{sec:more_method_details}. Then, we present more experiment results in Sec.~\ref{sec:more_exp_results}. Finally, we show more image, video, open-vocabulary, and interactive segmentation demos in Sec.~\ref{sec:more_vis_exp}.


\section{More Method Details}
\label{sec:more_method_details}

\noindent
\textbf{More Detailed Comparison with Recent Works.} Due to the page limitation, we only select several representative works for setting comparison.
Compared with specific models~\cite{qiao2021detectors, cheng2020panoptic}, our method achieves extreme parameter sharing and performs various tasks that these models cannot perform.

Compared with video segmentation and unified video segmentation~\cite{athar2023tarvis,li2023tube}, our method can also achieve open-vocabulary and interactive segmentation, as well as good enough performance on image segmentation. This is because our model is jointly co-trained on both image and video segmentation datasets without introducing task-specific tuning on video segmentation datasets. In addition, due to the frozen CLIP backbone, our method can also perform video open vocabulary segmentation without any architecture modification. 

Compared with recent partial unified models, our method achieves all related visual segmentation in one model.
For example, compared with Semantic-SAM~\cite{li2023semanticsam}, our model can achieve both video segmentation (VIS, VSS, VPS) and open-vocabulary segmentation. Compared with UNINEXT~\cite{UNINEXT}, our method can perform interactive segmentation, panoptic segmentation (VPS, PS), and open-vocabulary segmentation. Compared with OneFormer~\cite{jain2023oneformer}, we can achieve video, open-vocabulary, and interactive segmentation. Compared with TarVS~\cite{athar2023tarvis}, we can keep image segmentation without specific fine-tuning. Compared with recent FreeSeg~\cite{qin2023freeseg}, we can achieve both video segmentation and interactive segmentation in one model.


\noindent
\textbf{Implementation Details of OMG-Seg.} We use balanced training for our model. In particular, for two different setting of Tab.2 and Tab.3 in the main paper, we balance each dataset sample according to the COCO dataset size. Then, we choose the same data augmentation as Mask2Former~\cite{cheng2021mask2former}. For the text embedding generation, we follow the standard open-vocabulary detection and segmentation setting~\cite{wu2023open,zhou2022detecting}. We generate multiple text prompts with the class names and keep the text embedding fixed for both training and inference. In this way, we can achieve multi-dataset and open-vocabulary segmentation.

\noindent
\textbf{More Detailed Inference Process.} Our model has various inference modes. For image segmentation on various datasets, we simply follow the Mask2Former to obtain the corresponding mask and labels. For video segmentation, we adopt simple query matching~\cite{li2022videoknet,huang2022minvis} without learning the extra tracking query embedding. We believe adding such components will improve the video segmentation. For open-vocabulary segmentation, we fuse the frozen CLIP visual scope and predicted scope to boost the novel class segmentation. For interactive segmentation, we mainly use the point prompts to evaluate despite the box prompts, which can also be used as SAM~\cite{kirillov2023segment}. Moreover, since our model adopts the frozen CLIP features, we can freely label the prompt-driven segmentation masks, where we can achieve open-vocabulary interactive segmentation. The GFlops of the main paper are calculated with $1200 \times 800$ by default.

\section{More Experiment Results}
\label{sec:more_exp_results}

\begin{table}[t!]
\centering
\caption{Results using ResNet50 backbone.}
\resizebox{0.95\linewidth}{!}{
\begin{tabular}{c|ccccccc}
\toprule[0.2em]
Method & Backbone & COCO-PS & VIPSeg-VPS & Youtube-VIS-2019 \\
\midrule
Mask2Former~\cite{cheng2021mask2former} & ResNe50 & 52.0 & - & - \\
Mask2Former-VIS~\cite{cheng2021mask2former_vis} & ResNe50  & - & - & 46.4 \\
\hline
OMG-Seg & ResNe50 & 49.9 & 42.3 & 46.0 \\
OMG-Seg & ConvNext-L & 54.5 & 50.5 & 56.2 \\
\bottomrule[0.2em]
\end{tabular}
}
\label{tab-ablation:results_using_r50_backbone}
\end{table}

\begin{table}[t!]
\centering
\caption{Results using ViT backbone.}
\resizebox{0.95\linewidth}{!}{
\begin{tabular}{c|ccccccc}
\toprule[0.2em]
Backbone & COCO-PS & Youtube-VIS-2019 & VIP-Seg \\
\midrule
ViT-L (frozen) &  34.5  & 23.2 & 34.5 \\
ViT-L (learned) & 52.2 & 54.3 & 48.2 \\
ConvNext-L (frozen) & 54.5  & 56.2 & 50.5 \\
\bottomrule[0.2em]
\end{tabular}
}
\label{tab-ablation:results_using_vit_backbone}
\end{table}

\begin{table}[t!]
\centering
\caption{Ablation on self-attention mode for interactive segmentation tasks. We use ResNet50 as the backbone. The masks filter out the correlation of each query during self-attention.}
\resizebox{0.95\linewidth}{!}{
\begin{tabular}{c|ccccccc}
\toprule[0.2em]
Setting & COCO-PS & COCO-SAM  \\
\midrule
 Self Attention without masks  & 45.2 & 40.7 \\
 Self Attention with masks & 49.9 &  52.2 \\
\bottomrule[0.2em]
\end{tabular}
}
\label{tab-ablation:self_attention_masks_interative_mode}
\end{table}

In addition to the main paper, we also provide more ablation studies and experiment results here.

\noindent
\textbf{Results Using ResNe50 backbone.} In Tab.~\ref{tab-ablation:results_using_r50_backbone}, we report our model using ResNet50 backbone. We jointly co-train our model with 24 epochs. Compared with specific Mask2Former for 50 epoch training, our model can achieve considerable results but with less parameter costs.

\noindent
\textbf{Exploration on ViT-based CLIP backbone.} In Tab.~\ref{tab-ablation:results_using_vit_backbone}, we explore the CLIP-ViT backbone. We find using frozen CLIP-ViT leads to inferior results. This is because the position embedding of ViT is fixed (224 by default), and a simple bilinear upsampling operation hurts the origin representation. Thus, in the second row, we adopt the learned architecture. However, we still find performance gaps with convolution-based CLIP. Moreover, since there is no frozen CLIP and the open-vocabulary ability is lost during the fine-tuning. 


\noindent
\textbf{Interactive Segmentation with Masked Self-Attention.} In interactive mode, we set the query invisible (achieve this by masking) to each other during the cross-attention process. If not, as shown in Tab.~\ref{tab-ablation:self_attention_masks_interative_mode}, we find a significant performance drop for both COCO-SAM and COCO-PS. This is because, for interactive segmentation, the local features are good enough, while introducing the global information will bring noise to the query learning.


\begin{figure*}
    \centering
    \includegraphics[width=0.95\textwidth]{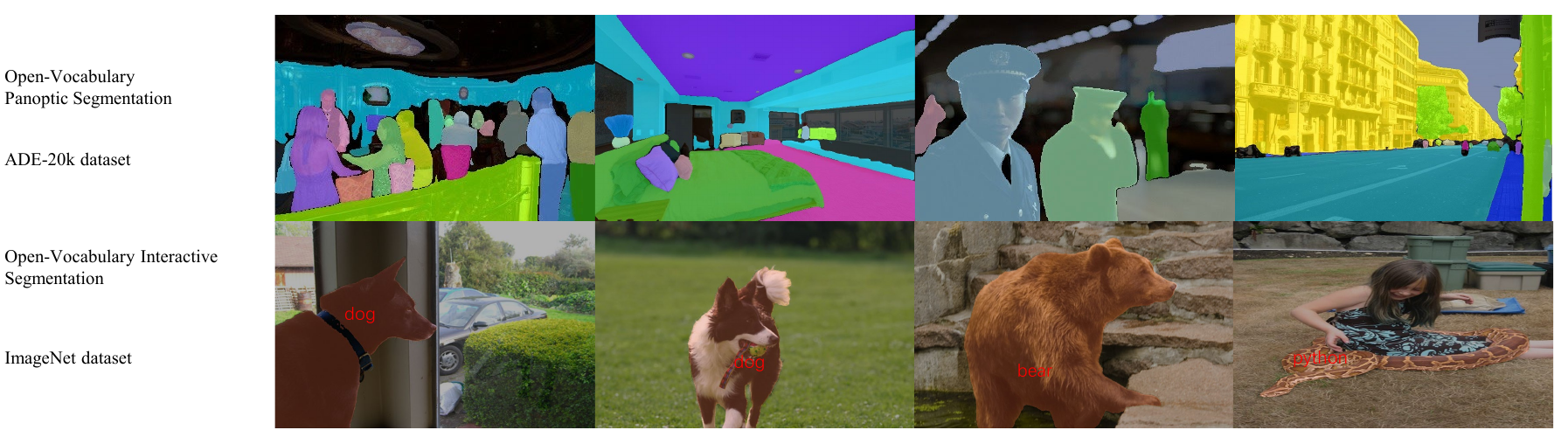}
    \caption{More functional Visualization of OMG-Seg model. In addition to five different tasks of the main paper, we also visualize the open-vocabulary segmentation results: open-vocabulary panoptic segmentation results on ADE-20k, open-vocabulary interactive segmentation results on ImageNet 1k dataset.}
    \label{fig:visual_res_supp}
\end{figure*}

\section{More Visualization Example}
\label{sec:more_vis_exp}



\noindent
\textbf{More Visual Results on More Tasks.} In Fig.~\ref{fig:visual_res_supp}, we present more visual examples for two additional tasks. One is open-vocabulary panoptic segmentation on ADE-20k. As shown in the top row, our method can achieve good zero-shot segmentation quality. In the second row, we also provide interactive segmentation on the ImageNet-1k dataset. We add the class labels that are from the simple CLIP score. To this end, we achieve open-vocabulary interactive segmentation.

\noindent
\textbf{Limitation and Future Work.} One limitation of our work is the capacity of our model. Since we use the frozen architecture to keep the open-vocabulary ability, which leads to inferior results for one specific dataset or task. However, we believe adding more dataset co-training~\cite{kirillov2023segment} with the learned backbone will improve our model performance. With the aid of more text-image pairs or classification datasets, we also achieve open-vocabulary segmentation ability while keeping the performance improved on close sets. This is our future work to scale up our model. Moreover, we can also add a text path to support language-driven segmentation tasks, such as referring image/video segmentation or even with large language models (LLMs) to perform joint reasoning and segmentation in one framework.